\documentclass[twoside,leqno,twocolumn]{article}  
\usepackage{ltexpprt} 

\usepackage{hyperref}
\RequirePackage[numbers]{natbib}

\usepackage{epsfig}
\usepackage{amsmath}
\usepackage{amsfonts}
\usepackage{amssymb}
\usepackage{framed}
\usepackage[normalem]{ulem}

\usepackage{jhh-misc-sdm}

\usepackage{graphicx}
\usepackage{caption}
\usepackage{subcaption}

\newcommand{\SX}{\mathscr S (\mathcal X)}
\newcommand{\SSX}{\mathscr S^{2}(\mathcal X)}

\def\[#1\]{\begin{align}#1\end{align}}
\def\(#1\){\begin{align*}#1\end{align*}}

\begin{document}

\title{\Large A Statistical Learning Theory Framework for Supervised Pattern Discovery} 

\author{Jonathan H. Huggins\thanks{MIT Department of EECS and CSAIL (\href{mailto:jhuggins@mit.edu}{jhuggins@mit.edu})}
\and
Cynthia Rudin\thanks{MIT Sloan School of Management and CSAIL (\href{mailto:rudin@mit.edu}{rudin@mit.edu})}}
	
\date{}

\maketitle

\begin{abstract} 
\small\baselineskip=9pt
This paper formalizes a latent variable inference problem we call {\em supervised pattern discovery}, the goal of which is to find sets of observations that belong to a single ``pattern.'' We discuss two versions of the problem and prove uniform risk bounds for both. In the first version, collections of patterns can be generated in an arbitrary manner and the data consist of multiple labeled collections. In the second version, the patterns are assumed to be generated independently by identically distributed processes. These processes are allowed to take an arbitrary form, so observations within a pattern are not in general independent of each other. The bounds for the second version of the problem are stated in terms of a new complexity measure, the quasi-Rademacher complexity. 
\end{abstract}

\section{Introduction}

The problem of \textit{supervised pattern discovery} is that of finding sets of observations that belong together, given a set of past patterns to learn from. This problem arises naturally in domains ranging from computer vision to crime data mining. We provide a formal definition and theoretical foundation for pattern discovery. Unlike the classical problem of classification, in pattern discovery we posit an infinite number of classes (``patterns"), each containing a finite number of observations. Also, in contrast to standard classification assumptions, we do not expect the observations to be chosen i.i.d. On the contrary, the observations within a pattern may be highly correlated, whereas the latent patterns are chosen i.i.d., and our goal is to locate these patterns among the full set of observations.
We briefly outline three motivating examples. 

The first is a problem that is faced every day by crime analysts in police departments across the world \citep{Chen:2004,Nath06crimedm}. These analysts spend much of their time searching for patterns of crimes within databases of crime reports. The identification of emerging patterns of crime has been a key priority of the crime analysis profession since its inception. Analysts have knowledge of past patterns of crime, which they generalize to detect new patterns in recent crimes. Each pattern thus corresponds to the (group of) people committing the crimes. Since there are an unknown number of criminals in the world, and as patterns of crimes happen in myriad ways, we cannot assume a fixed, finite number of patterns. In addition, crimes committed by the same individual are certainly not i.i.d. On the other hand, the patterns are similar enough that sometimes the analysts can identify them. Thus, instead of the usual i.i.d.~assumption of observations made generally in machine learning, we might consider each observation (crime) as being generated by one of many latent processes (the criminals), chosen i.i.d. Observations generated by the same process are considered part of a single pattern, and all of the observations are visible simultaneously. Automated methods for crime pattern detection include a neural network approach \citep{Dahbur:2003} and a greedy pattern-building algorithm \citep{Wang:2013}. Others \citep{Lin03anoutlier-based,brown2003data} investigate the slightly easier task of finding pairs of related crimes. 

\begin{figure*}[tbp] 
\centering
\begin{subfigure}[b]{.3\textwidth}
	\centering
	\frame{\includegraphics[width=\textwidth]{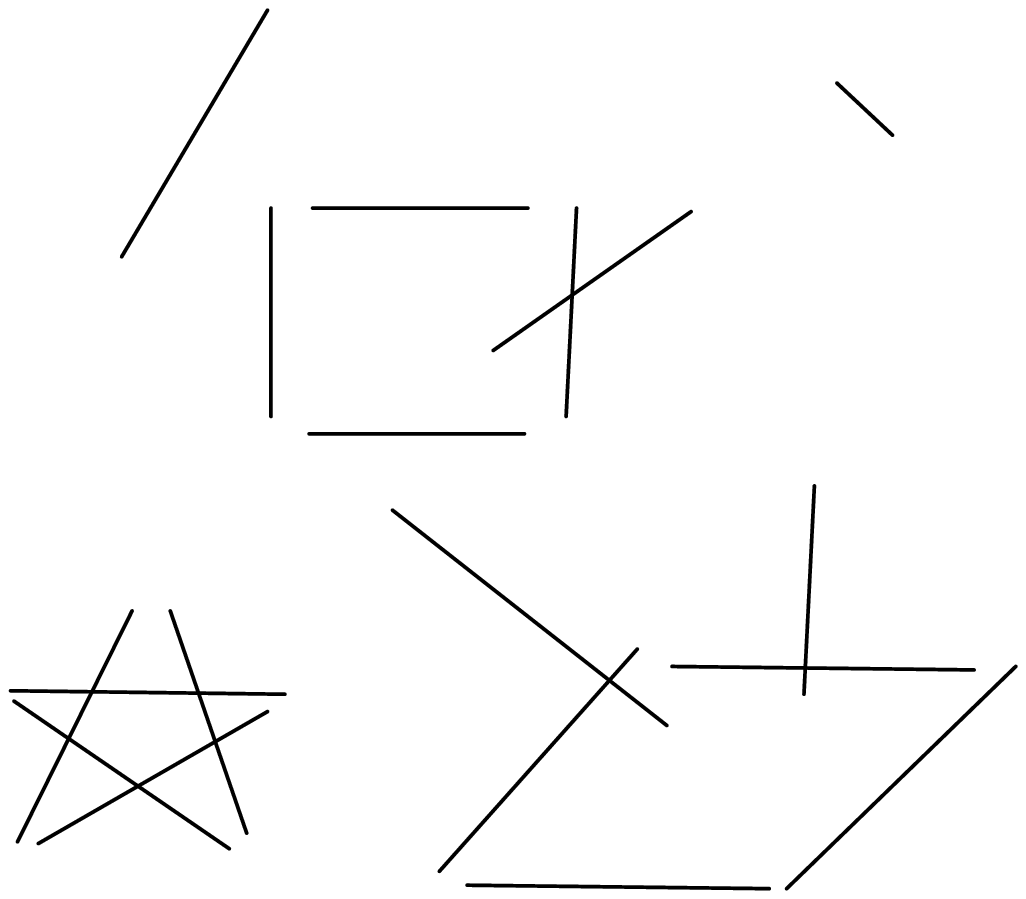}}
	\caption{Three simple patterns}
	\label{subfig:simple-patterns}
\end{subfigure}
\qquad\qquad
\begin{subfigure}[b]{.3\textwidth}
	\centering
	\frame{\includegraphics[width=\textwidth]{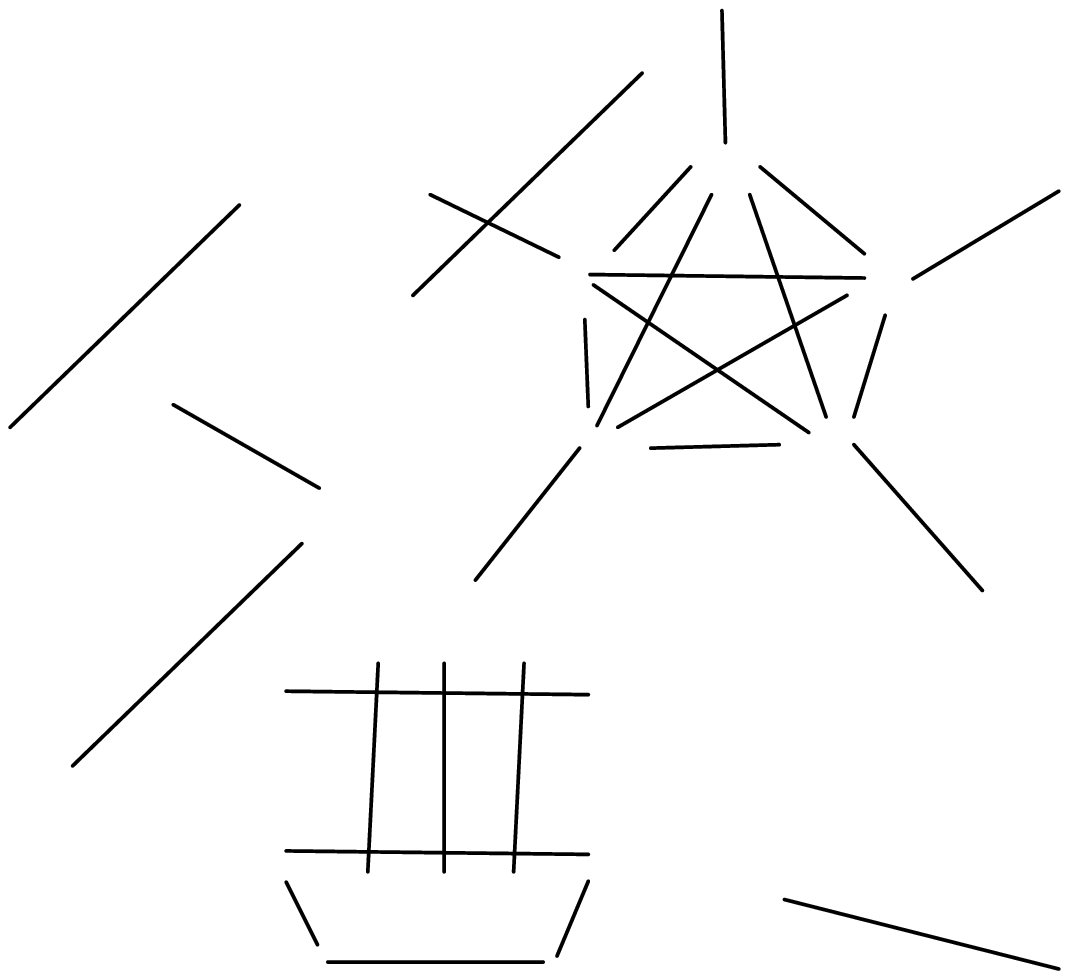}}
	\caption{Two more complex patterns}
	\label{subfig:complex-patterns}
\end{subfigure}
\caption{Examples of block patterns using lines, with extra lines that do not belong to a multi-line pattern.} 
\label{fig:block-patterns}
\end{figure*}

As a second example, consider the following simplified perceptual organization-style problem,\footnote{Ideas from perceptual organization have proven to be very useful in the field of computer vision \citep{Sarkar:2003}.} which involves finding geometric patterns in an image (cf. \cite{Ferrari:2006,Zhu:2008}, and also \cite{Payet:2010} for an unsupervised variant). Each observation is a line segment in $\reals^{2}$. A robot or human might observe an image with more than one pattern in it: say, a star, a square, and a rhombus (see Figure \ref{subfig:simple-patterns}), which are placed according to a particular probability distribution within the space. The goal is to find the patterns, where each pattern consists of a subset of observations. In this case, a single observation can only be classified in the context of the other observations. We do not know in advance what constitutes a pattern; we have only a labeled set of other patterns to learn from. There may actually be an infinite number of pattern types. For instance, while a human might quickly recognize the two patterns in Figures \ref{subfig:complex-patterns}, there are an infinite number of other possible patterns they might also recognize in some other collection of line segments. 

A final example that falls within the pattern discovery framework comes from personalized medicine \citep{Ginsburg:2001,Hamburg:2010}. In personalized medicine, an individual's molecular and genetic profile is used to develop a specialized treatment for that person. To accomplish this, patterns must be found within individuals' molecular/genetic profiles, the progressions of their symptoms, and the results of their treatments. These patterns are used  not just to decide between one or two possible treatments. Instead, a large number of treatment regimens may be discovered, with each regimen potentially applying to only a small number of patients. For example, personalized medicine approaches have found particular success in using genome-wide gene-expression data  for the treatment of cancer \citep{Sorlie:2001,vantVeer:2002,Wang:2011}.
``Cancer'' is a highly amorphous term. While certain cancers are caused by a few well-understood gene mutations, in many cancers there are a large number of infrequent mutations that each make a small contribution to tumorigenesis \citep{Vogelstein:2013}. For example, breast cancer is caused by hundreds if not thousands of different mutations, with only three point mutations and perhaps ten recurrent mutations occurring in more than 10\% of cases \citep{Kobolt:2012}. Thus, flexible pattern discovery methods are required \citep{Sorlie:2001,vantVeer:2002}. For a range of personalized medicine examples and references, see the proceedings of the recent NIPS 2010 Workshop on Predictive Models in Personalized Medicine\footnote{\url{https://sites.google.com/site/personalmedmodels/}}. 

In this paper we develop a statistical learning theory framework for two versions of the problem of supervised pattern discovery, providing a theoretical foundation for applications that are already used in practice for pattern discovery. In particular, we develop uniform risk bounds that can be used for empirical risk minimization \citep{Vapnik:1998}. We call the first version of the pattern discovery problem \textit{block pattern discovery}. The block problem assumes there are collections of patterns. The collections are i.i.d.\ but the pattern-generating mechanisms within each collection are not necessarily independent. The second version is the {\em individual pattern problem}, in which the patterns are presented as a single collection and the pattern generating processes (with each generating a single pattern) are i.i.d.

To our knowledge, there are no other learning theory frameworks which, like pattern discovery, allow for an infinite number of latent patterns, each with a finite number of observations. Statistical learning theory for classification \citep{Vapnik:1998} supposes a finite number of possible classes each containing an infinite number of observations in the limit of infinite data, and many other supervised problems (e.g.\ supervised ranking) are similar. Supervised clustering \citep{Balcan:2008,Awasthi:2010} similarly posits a known, finite number of clusters to which observations in some fixed data set belong. In the clustering model there is a ``teacher'' who provides feedback about the correctness of the proposed clustering and the clustering rule is assumed to come from some known concept class. Algorithms operating within the framework are concerned with finding the true rule using a polynomial number of queries to the teacher. Finally, standard clustering is an {\em unsupervised} method \citep{HTF:2008}, whereas pattern discovery is concerned with {\em supervised learning}.  

In addition to the theoretical work just described, there is a large body of work in the statistics, machine learning, and data mining communities on clustering (mixture modeling) with an infinite number of clusters (components). A popular approach to infinite mixture modeling is based on Bayesian non-parametric models, particularly the Dirichlet process (DP). Both unsupervised \citep{Rasmussen:1999,Neal:2000}
and semi-supervised \citep{Akova:2012} approaches have been developed. Note, however, that DP-based approaches produce clusters of infinite size in the infinite data limit, whereas we shall be interested in clusters that are of finite size even in the infinite data limit. 

The remainder of the paper is organized as follows. In Section \ref{sec:block}, notation is established and the block pattern discovery problem is defined. In Section \ref{sec:block-bounds}, we give risk bounds for the block problem in terms of covering numbers. The individual pattern discovery problem is defined in Section \ref{sec:individual} while Section \ref{sec:individual-bounds} gives risk bounds for the individual pattern discovery problem based on an adaptation of the Rademacher complexity measure that is appropriate for pattern discovery.

\section{The Block Pattern Discovery Problem} \label{sec:block}

We first investigate what we call the {\em block pattern discovery} problem. The observations are a sequence of i.i.d.\  {\em blocks}, where a {\em block} is a collection of patterns. For example, a single block could be images with lines that form patterns, as in Figures \ref{fig:block-patterns}a and \ref{fig:block-patterns}b. In order words, each line would be an observation, a set of lines would form a geometric pattern, and the image containing the geometric patterns would constitute a block. The goal is then to find the patterns in new groups of observations, such as in new images. 

Let $\mcX$ be the observation space and define $\SX$ to be the set of finite, non-empty subsets of $\mathcal X$. A pattern will consist of one or more observations from $\mcX$, so $\SX$ defines the collection of all possible patterns that could be observed. Also, let $\SSX := \mathscr S(\SX)$ denote all finite collections of patterns. In other words, $x \in \mcX$ is a single observation; $P \in \SX$ is a pattern, which consists of a finite number of observations; and $Q \in \SSX$ is a finite collection of patterns (i.e.\ $Q = \{P_{1},P_{2},\dots, P_{k}\}, P_{i} \in \SX$). For example, in the crime pattern detection application, $x \in \mcX$ is a single crime, $P_{i} \in \SX$ are the crime patterns---crimes committed by a single person or group---and $Q = \{P_{1},P_{2},\dots, P_{k}\} \in \SSX$ is the collection of crime patterns. Note that each observation can belong to only one pattern.

Define an (unknown) distribution $\mcD$ over collections of patterns, so $Q \sim \mcD$ is an element of $\SSX$ and can be written as $Q = \{ P_{1},\dots,P_{k} \}$, where each $P_{i} \in \SX$ is a single pattern. Note that $k$ is itself random. $Q$ can be thought of as representing a labeled version of the observations --- that is, indicating which observations are part of the same pattern. Let $X_{Q} = \bigcup_{P \in Q} P \in \SX$ be the set of observations associated with $Q$, which would correspond to an unlabeled dataset with latent patterns defined by $Q$. 

Let $S : \SSX \to \SSX$ be a {\em selector function}, which maps each collection of patterns $Q \in \SSX$ to a subset of $\mathscr S( X_{Q})$, the subsets of observations derived from $Q$. The function $S(Q)$ is used to select out which subsets of the set of observations $X_{Q}$ the loss function will depend on. Since $S$ is a function of $Q$, these subsets can depend on the true patterns in the observations. We will be interested in choosing a {\em pattern discovery function} $f : \SX \times \SX \to [0,1]$. The function $f(X, U)$ outputs a score between 0 and 1, where 1 (resp. 0) indicates complete confidence that $U$ is part (resp. not part) of a pattern from $X$. Let $\mcF$ be a family of pattern discovery functions. We assume throughout that for $f \in \mcF$, $f(X, U) = 0$ if $U \not\subseteq X$, since in this case it is obvious that $U$ cannot be part of a pattern from $X$. 

The {\em block loss functional} $\mcL_{S}$ under selector $S$ measures the performance of the pattern discovery function $f \in \mcF$ on a collection of patterns $Q \in \SSX$ and is defined to be
\begin{align}
\mcL_{S}(f; Q) =
\left[\frac{1}{Z_{Q,S}}\sum_{U \in S(Q)} \ell_{sp}^{2}(f, U, Q) \right]^{1/2}, 
\end{align}
where $Z_{Q,S} = |S(Q)|$ is the normalization function so $\mathcal L_{S} \in [0,1]$ and $\ell_{sp}(f, U, Q) \in [0,1]$ is the {\em local loss} of $f$ on a subset $U$ when the true pattern collection is $Q$. In the following section, we will focus on the case where $\ell_{sp}(f, U, Q) = |\mathbb I(U \subset P \in Q) - f(X_{Q}, U)|$, which penalizes $f$ for how far it is from the indicator of whether $U$ is part of a pattern. 

One particular case of interest is when the selector function is $A : Q \mapsto \mathscr S(X_{Q})$, which selects all subsets of the data. This maximal selector can be thought of as the ideal one, in the sense that $\mcL_{A}$ considers the performance of $f$ on all possible subsets. However, since there are an exponential number of subsets, evaluating $\mcL_{A}$ is not usually practical. In such cases, a selector that picks out subsets deemed ``important'' could be used. For example, the selector might choose all subsets of true patterns as positive examples and subsets of true patterns with one additional data point not from that pattern as negative examples. This particular selector function is useful for training greedy algorithms that build patterns incrementally. These kinds of greedy algorithms for finding patterns have been used both in supervised settings (e.g.\ that of \citep{Wang:2013}) and unsupervised settings (e.g.\ work on set expansion \citep{WangCo2008,LethamRuHe13}).



\section{Risk Bounds for the Block Problem} \label{sec:block-bounds}

Let the true risk for $f \in \mcF$ be 
\beqn R(f) = \bbE_{Q \sim \mcD} \mathcal L_{S}(f; Q) \eeqn
and the empirical risk for $f$ given $Q_{1},\dots,Q_{n} \overset{i.i.d.}{\sim} \mcD$ be
\beqn \hR_{n}(f) = \frac{1}{n} \sum_{i=1}^{n}\mathcal L_{S}(f; Q_{i}). \eeqn
We develop 
bounds on the difference $L_{n}(f) = R(f) - \hR_{n}(f)$ between the true and empirical risk. These bounds adapt and expand on classical learning theory results to the new pattern detection problem. Indeed, our main goal in this section is to point out that certain pattern discovery problems can be framed such that they inherit standard i.i.d.~learning theory guarantees.

Because we have now packaged the block discovery problem into a learning theoretic framework, we can apply Rademacher complexity bounds, which we then relate to empirical $\ell_{2}$ covering numbers via Dudley's entropy bound. Define the loss class to be $\mcG := \mcL_{S} \circ \mcF := \{ g = \mcL_{S}(f; \cdot) \given f \in \mcF \}$, and  $\tilde\mcG = \{ Q \mapsto \mcL_{S}(f; Q) - \mcL_{S}(0; Q) \given f \in \mcF\}$ to be the offset loss class. Also define the empirical metric
\beqn d_{n}(g_{1}, g_{2}) =  \left[n^{-1}\sum_{i=1}^{n} (g_{1}(Q_{i}) - g_{2}(Q_{i}))^{2}\right]^{1/2}. \eeqn

\bdefn For metric space $(T, d)$, the {\bf $\eps$-covering number} $\mcN(T, d, \eps)$ is defined to be the smallest integer $K$ such that there are points $x_{1},\dots,x_{K} \in T$ satisfying $\bigcup_{i=1}^{K}B_{\eps}(x_{i}) \subseteq T$, where $B_{\eps}(x_{i})$ is the open ball in $T$ of radius $\eps$ centered at $x_{i}$. 
\edefn

\bthm \label{cor:l2-bound-g-metric} 
 Let $Q_{1},\dots,Q_{n} \overset{i.i.d.}{\sim} \mcD$. Then for any positive integer $n$ and any $0 < \delta < 1$, with probability $1 - \delta$ over samples of length $n$, every $f \in \mcF$ satisfies
\begin{align}
L_{n}(f) &\le \bbE  \int 24\sqrt{\frac{\log \mcN(\tilde\mcG, d_{n}, \eps)}{n}}\,d\eps + \sqrt{\frac{8 \ln (2/\delta)}{n}}. \label{eq:l2-bound-g}
\end{align}
\ethm
\bprf
We combine a Rademacher complexity bound with Dudley's entropy bound, in the Supplemental Material.
\eprf

We next derive a risk bound in terms the covering number for the underlying class $\mcF$ of pattern discovery functions. In some sense this is a more natural covering number to consider than the covering number of the shifted loss class $\tilde\mcG$ used in Theorem \ref{cor:l2-bound-g-metric}, since the ultimate goal is to choose a function from $\mcF$, not $\tilde\mcG$. 
To obtain a relationship between covering numbers for $\mcF$ and $\tilde\mcG$, we must define a metric on $\mcF$ and relate it to the metric on $\tilde\mcG$. First, for $f_{1}, f_{2} \in \mcF$, define the (squared) metric for a single collection to be (with $Z := Z_{Q,S}$)
\beq \ell_{Q}^{2}(f_{1},f_{2}) := \frac{1}{Z}\sum_{U \in S(Q)} [f_{1}(X_{Q}, U) - f_{2}(X_{Q}, U)]^{2}. \eeq
The empirical metric we are interested in is 
\beq \ell_{n}(f_{1},f_{2}) := \left[n^{-1}\sum_{i=1}^{n} \ell_{Q_{i}}^{2}(f_{1}, f_{2})\right]^{1/2}. \eeq
\bthm \label{thm:l2-bound-f-metric}
Under the same hypotheses as Theorem \ref{cor:l2-bound-g-metric},
\begin{align*}
L_{n}(f) &\le \bbE \int 24 \sqrt{\frac{\log \mcN(\mcF, \ell_{n}, \eps)}{n}}\,d\eps +  \sqrt{\frac{8 \ln (2/\delta)}{n}}. 
\end{align*}
\ethm
\bprf
See the Supplementary Material. 
\eprf
Hence, we can perform empirical risk minimization in the block pattern discovery framework.

\section{The Individual Pattern Discovery Problem} \label{sec:individual}

The block formulation of the pattern discovery problem applies to the examples from vision, crime, and medicine outlined in the introduction, but in many cases the structure of the problem may be different. Instead of working from blocks of patterns, and learning across blocks, we might wish to learn across patterns within a single block. 
In the crime example, each pattern collection $Q_{i}$ represents an entire set of crimes, perhaps from several cities or different intervals of time. For a police department wanting to evaluate their pattern detection ability on individual patterns of crime, rather than crimes within blocks, we should try to predict when only a single pattern collection $Q$ is available. 

In the {\em individual pattern  discovery} problem, the learner must use one collection $Q$ as training data instead of multiple collections $Q_{1},\dots,Q_{n}$. The task is then to partition newly observed data $X \in \SX$ into patterns.  We assume the finite patterns arise from stochastic processes that are chosen i.i.d.~from an unknown probability distribution over processes. We allow these processes to take an arbitrary form (since, for example, a single criminal's crimes certainly are not i.i.d.). The task is to identify patterns in the observations that are not labeled as such. The individual pattern problem can thus be viewed as a kind of supervised, latent variable problem, where the pattern generating processes are the latent variables. 

To formally define the individual pattern problem, let $\mcP$ be a distribution over patterns, so if $P \sim \mcP$, then $P \in \SX$. We are given data $P_{1},\dots,P_{n} \overset{i.i.d.}{\sim} \mcP$, which together form a collection of patterns $Q = \{ P_{1},\dots,P_{n}\}$. Note that $i$ now indexes over patterns and $n$ denotes the number of patterns, not the number of pattern collections. Although the processes could themselves be random, since we assume all observations from each process are part of $Q$, all of the randomness of the processes can be absorbed into $\mcP$. It is therefore safe to equate each process with the pattern it generates.  

As in the block case, we wish to choose a pattern discovery function $f : \SX \to [0,1] \in \mcF$ to minimize a loss functional, though now $f$ does not have access to the whole set of observations $X_{Q}$, only the subset $U \subset X_{Q}$ that it is making a decision on. Writing $X$ in place of $X_Q$ when the underlying partition is unknown, the loss function is defined to be:
\[
\mathcal L_{\alpha}(f; P, X) &:= \frac{\alpha}{Z_{P}}\sum_{U \in \mathscr S(P)} \ell_{+}(f, U)   \label{eq:L-alpha} \\
& + \frac{1 - \alpha}{Z_{P,X}}\sum_{U \in S_{-}(P, X)} \ell_{-}(f, U), \nonumber
\]
where $Z_{P} = |\mathscr S(P)|$ and $Z_{P,X} = |S_{-}(P,X)|$ are normalization functions.
The functionals $\ell_{+}, \ell_{-} \in [0,1]$ define the losses on positive and negative examples, respectively.  $S_{-}(P,X)$ is the {\em negative example selector function}, which plays an analogous role to selector function for the block loss functional.  The weight factor $0 < \alpha < 1$ determines the relative importance of positive examples compared to negative examples, making the loss cost sensitive. It is necessary to weight the two sums, since otherwise in the limit as $n \to \infty$, the value of the loss could be determined solely by the negative examples. This is because as $n \to \infty$, $|X| \to \infty$ and thus $|S_{-}(P,X)| \to \infty$ while $|\mathscr S(P)|$ remains finite.

As with the general selector function $S$ for the block loss, choosing $S_{-}$ to select all negative examples would involve an exponentially large (in $|X|$) number of subsets. Therefore, instead we might define
\begin{equation}
S_{-}(P,X) = \{ U \cup \{ x \} \given U \in \mathscr S(P), x \in X \setminus P \}. \label{eq:negative-selector}
\end{equation}
That is, we look at how $f$ performs on sets that are almost patterns. This choice of selector is particularly relevant for greedy algorithms, and is used by \citet{Wang:2013}. We will assume that  $S_{-}$ takes the  form of equation \eqref{eq:negative-selector}, though our results can easily be adapted to other choices of $S_{-}$ that treat the elements of $X$ uniformly. 

\section{Risk Bounds for the Individual Pattern Discovery Problem}
\label{sec:individual-bounds}

We will prove two results inspired by \citet{Bartlett:2002}. We will first introduce a new Rademacher complexity-like quantity to use in place of the covering number term. We will also show this quantity can be well-estimated empirically. As before, we define the empirical risk 
\beqn \hR_{\alpha,n}(f) := n^{-1} \sum_{j=1}^{n} \mcL_{\alpha}(f, P_{j}, X_{Q}), \eeqn
the true risk $R_{\alpha,n}(f) := \bbE \hR_{\alpha,n}(f)$, and denote the difference by $L_{\alpha,n}(f) := \hR_{\alpha,n}(f) - R_{\alpha,n}(f)$. Note that unlike with the empirical risk $\hR_{n}$ in the block problem, the terms in the sum defining $\hR_{\alpha,n}(f)$ are not independent since they are all a function of $X_{Q}$. Recall that $\hR_{n}(f) = \frac{1}{n} \sum_{i=1}^{n}\mathcal L_{S}(f; Q_{i})$, where the $Q_{i}$'s are all independent of each other. Let
\begin{align}
\lefteqn{\hat\mcQ_{n}(\mcL_{\alpha}, \mcF)}   \\
&:= \bbE_{\veps}\bigg[\sup_{f \in \mcF} \big| n^{-1}\textstyle\sum_{i=1}^{n} \veps_{i} \mcL_{\alpha}(f, P_{i}, X_{Q})\big| \bigg| Q\bigg], \nonumber
\end{align}
where the $\veps_{i}$ are independent uniform $\{\pm 1\}$-valued random variables. Define the {\em quasi-Rademacher complexity} to be 
\beqn
\mcQ_{n}(\mcL_{\alpha}, \mcF) := \bbE_{Q} \hat\mcQ_{n}(\mcL_{\alpha}, \mcF).
\eeqn
The quasi-Rademacher complexity is distinct from standard Rademacher complexity because the terms in the sum defining $\hat\mcQ_{n}(\mcL_{\alpha}, \mcF)$ are dependent via $X_{Q}$. Hence, like Rademacher complexity, it measures the ability of the function class $\mcF$ to fit random noise, though unlike Rademacher complexity, the loss function depends on all the observations. 

The two main results in this section are based on McDiarmid's inequality  \citep{McDiarmid:1998}. For independent random variables $Y_{1},\dots,Y_{n}$ taking values in a set $V$, assume that the function $F : V^{n} \to \reals$ satisfies the condition that, for all $1 \le i \le n$,
\beqn
\sup_{y_{1},\dots,y_{n},y_{i}' \in V} |F(y_{1},\dots,y_{n}) - F(y_{1},\dots,y_{i}',\dots,y_{n})| \le c. 
\eeqn
In ``expectation form,'' McDiarmid's inequality states that for any $0 < \delta < 1$, with probability at least $1 - \delta$, 
\beqn
F(Y_{1},\dots,Y_{n}) \le \bbE F(Y_{1},\dots,Y_{n}) + \sqrt{\frac{n c^{2}\log(1/\delta)}{2}}.
\eeqn
%

In order to apply McDiarmid's inequality, we require the following lemma. 
\blem \label{lem:loss-change}
Assuming $|P_{i}| \le B$ almost surely, then if one $P_{i}$ changes, the value of $\hR_{\alpha,n}(f)$ changes by at most $B_{\alpha}/n = \left(1 + 2(1-\alpha)B\right)/n$.  
\elem
\bprf See the Supplemental Material. \eprf

To ensure generalization, we impose a constraint on the distribution of the number of observations $|P|$ in a pattern $P \sim \mcP$. The first result assumes that $|P|$ is bounded.
\bthm \label{thm:quasi-rademacher-bound}
Let $P_{1},\dots,P_{n} \overset{i.i.d.}{\sim} \mcP$ and assume $|P_{i}| \le B$ almost surely. Then for any positive integer $n$ and any $0 < \delta < 1$, with probability $1 - \delta$ over samples of length $n$, every $f \in \mcF$ satisfies
\beqn L_{\alpha,n}(f) \le 2\mcQ_{n}(\tilde\mcL_{\alpha}, \mcF) + \sqrt{\frac{8B_{\alpha}^{2} \ln (2/\delta)}{n}}, \eeqn
where $B_{\alpha} = 1 + 2(1-\alpha)B$ and $\tilde\mcL_{\alpha}(f; \cdot, \cdot) := \mcL_{\alpha}(f; \cdot, \cdot) - \mcL_{\alpha}(0; \cdot, \cdot)$ is the shifted loss. 
\ethm
\bprf 
See the Supplemental Material. 
\eprf

Even though all of the $\mcL_{\alpha}$ terms in the risk are related through $X_{Q}$, we are able to control the effect of changing one $P_{i}$ on the losses $\mcL_{\alpha}(f; P_{j}, X_{Q})$ when $j \ne i$. This is the key to being able to design a bound for problems as complex as supervised pattern detection. If the distribution on $|P_{i}|$ is arbitrary, it is not in general possible to obtain bounds such as the one above. However, we can relax the assumption that the size is bounded and instead assume geometric tails for $|P_{i}|$. Under this weaker condition we achieve an $O(\sqrt{\log n / n})$ convergence  rate instead of $O(1 / \sqrt n)$.
\bthm \label{thm:quasi-rademacher-geom-bound}
Let $P_{1},\dots,P_{n} \overset{i.i.d.}{\sim} \mcP$ and assume that there exists a natural number $B_{0}$, a constant $C$, and a rate $0 < \lambda < 1$ such that for any $B \ge B_{0}$,  $\Pr[|P_{i}| > B] \le C\lambda^{B}$. Furthermore, assume that 
\beqn B_{n} := \left\lceil\frac{\log(2Cn/\delta)}{\log(1/\lambda)}\right\rceil  \ge B_{0}. \label{eq:B-for-geom-bound} \eeqn
Then for any positive integer $n$ for which equation \eqref{eq:B-for-geom-bound} holds and all $0 < \delta < 1$, with probability $1 - \delta$ over samples of length $n$, every $f \in \mcF$ satisfies
\beqn L_{\alpha,n}(f) \le 2\mcQ_{n}(\tilde\mcL_{\alpha}, \mcF) + \sqrt{\frac{8B_{n,\alpha}^{2} \log(4/\delta)}{n}}, \eeqn
where
\beqn B_{n,\alpha} = 1 + 2(1-\alpha)B_{n}. \label{eq:B-n-alpha} \eeqn
\ethm
\bprf If we choose some fixed $B \ge B_{0}$, then consider the probability that the size of all patterns is at most $B$,
\beqa
\Pr[|P_{i}| \le B~\forall i = 1,\dots,n] &=& (1 - \Pr[|P_{i}| > B])^{n} \\
&\ge& 1 - n \Pr[|P_{i}| > B] \\
&\ge& 1 - nC\lambda^{B},
\eeqa
so the hypotheses required for Theorem \ref{thm:quasi-rademacher-bound} hold with probability at least $1 - nC\lambda^{B}$. If we set $nC\lambda^{B} \le \delta/2$ and make the substitution $\delta \to \delta/2$ in the statement of Theorem  \ref{thm:quasi-rademacher-bound}, by the union bound, with probability at least $1 - \delta$, every $f \in \mcF$ satisfies
\beq L_{\alpha,n}(f) \le 2\mcQ_{n}(\tilde\mcL_{\alpha}, \mcF) + \sqrt{\frac{8B_{\alpha}^{2} \ln (4/\delta)}{n}}. \eeq
Solving $nC\lambda^{B} \le \delta/2$ for $B$ implies that the minimal choice for $B$ is
\beq
B_{n} = \left\lceil\frac{\log(2Cn/\delta)}{\log(1/\lambda)}\right\rceil. 
\eeq
The result now follows by substituting $B_{n}$ for $B$ in the expression for $B_{\alpha}$. 
\eprf

{\em Remark 1.} The risk bound given in Theorem \ref{thm:quasi-rademacher-bound} shows that an important parameter in determining the difficulty of the pattern discovery problem is the size of the patterns. If the patterns $P_{1}, P_{2}, \dots$ contain a small number of elements (i.e., $B$ is small), then pattern discovery will be easier. We note that the cost paid for large $B$ is only linear in $B$. In the case of Theorem \ref{thm:quasi-rademacher-geom-bound} where patterns can be arbitrarily large, tighter risk bounds are possible when most of the patterns are small.

{\em Remark 2.} The theorems proven in this section are stated in terms of the number of patterns. However, risk bounds are typically given in terms of the number of observations made. While the bounds in terms of the number of patterns are tighter, they are, essentially, asymptotically equivalent to those stated in terms of the number of observations. To see this, first consider the case where $|P|$ is almost surely bounded by $B$ and let $m$ be the number of observations made. Then 
\beq m = \sum_{i=1}^{n} |P_{i}| \le nB \eeq
Hence, the bound in Theorem \ref{thm:quasi-rademacher-bound} can be rewritten in terms of $m$, sacrificing at most a factor of $\sqrt B$. 
\bcor  \label{cor:quasi-rademacher-bound}
Under the same hypotheses as Theorem \ref{thm:quasi-rademacher-bound}, if $m$ is the number of observations taken, then for any positive integer $n$ and any $0 < \delta < 1$, with probability $1 - \delta$ over samples of length $n$, every $f \in \mcF$ satisfies
\beqn L_{\alpha,n}(f) \le 2\mcQ_{n}(\tilde\mcL_{\alpha}, \mcF) + \sqrt{\frac{8B \cdot B_{\alpha}^{2} \ln (2/\delta)}{m}}.  \eeqn
\ecor

An analogous result for the case when $|P_{i}|$ has geometric tails can be state based on the following simple fact.
\blem \label{lem:obs-bound}
If $|P_{i}|$ has geometric tails, then the expected number of observations in a pattern is at most 
\beqn B_{\lambda,C} :=  B_{0} + \frac{C\lambda^{B_{0}+1}}{1 - \lambda}(B_{0} + 1/(1 -\lambda)). \eeqn
\elem 
\bprf The proof is given in the Supplementary Material and follows by standard geometric series properties.
\eprf

Since the tails of $|P_{i}|$ are geometric, the probability of $|P_{i}|$ being much greater than $B_{\lambda,C}$ is exponentially small. Thus, with high probability, the bound in Theorem \ref{thm:quasi-rademacher-geom-bound}, restated in terms of $m$ (as in Corollary \ref{cor:quasi-rademacher-bound}), is only worsened by a factor of $O(\sqrt{B_{\lambda,C}})$. 

\subsection{Estimating $\mcQ_{n}$}
Like Rademacher complexity, quasi-Rademacher complexity can be empirically estimated efficiently. 

\bthm 
Assuming $|P_{i}| \le B_{0}$ almost surely~for $P_{i} \overset{i.i.d.}{\sim} \mcP$, then for any natural number $n$ and any $0 < \delta < 1$, with probability $1-\delta$ over $Q$ and $\veps$
\beqa 
\bigg| \mcQ_{n}(\tilde\mcL_{\alpha}, \mcF) -  \sup_{f \in \mcF} |\hat\mcQ_{n}(\tilde\mcL_{\alpha}, f)| \bigg| \le \sqrt{\frac{8B_{\alpha}^{2} \ln (2/\delta)}{n}} 
\eeqa
and with probability $1-\delta$ over $Q$
\beqa
\bigg| \mcQ_{n}(\tilde\mcL_{\alpha}, \mcF) - \hat\mcQ_{n}(\tilde\mcL_{\alpha}, \mcF) \bigg| \le \sqrt{\frac{8B_{\alpha}^{2} \ln (2/\delta)}{n}}
\eeqa
where 
\beq \hat\mcQ_{n}(\tilde\mcL_{\alpha}, f) := n^{-1}\textstyle\sum_{i=1}^{n} \veps_{i} \mcL_{\alpha}(f, P_{i}, X_{Q}). \eeq
\ethm
\bprf A result analogous to Lemma \ref{lem:loss-change} can be proven for $\hat\mcQ_{n}(\tilde\mcL_{\alpha}, f)$ in place of $\hat\mcR_{n}(f)$, since they are the same up to changes in signs of the terms induced by the $\veps_{j}$. Thus, the proof is essentially identical. The theorem then follows from McDiarmid's inequality applied as in the proof of Theorem \ref{thm:quasi-rademacher-bound}. \eprf
Similar bounds can be obtained in the case that $|P_{i}|$ has a geometric tail (as long as equation \eqref{eq:B-for-geom-bound} holds) with the right hand sides of the inequalities in the previous theorem replaced by 
\beq \sqrt{\frac{8 B_{n,\alpha}^{2} \log(4/\delta)}{n}}, \eeq
where $B_{n,\alpha}$ is given in Theorem \ref{thm:quasi-rademacher-geom-bound}. 

\section{Algorithms and Applications for Individual Pattern Discovery}\label{sec:algs}

The theoretical guarantees in Section \ref{sec:individual} lead directly to the following general algorithm for individual pattern discovery. Before running this algorithm, the user:
\begin{enumerate}
\item chooses a parametric class of pattern discovery functions $f_{\beta}:\SX\rightarrow [0,1]$, where $\beta \in \Gamma$; and 
\item chooses a threshold $\theta\in [0,1]$. If a subset of observations $\hat{P}$ scores below $\theta$, that is $f_{\beta}(\hat{P})\leq\theta$ then we would not consider $\hat{P}$ a pattern.
\end{enumerate}
The algorithm is as follows: \\
\rule{.45\textwidth}{.1pt}
\begin{algorithm} \label{Alg:Ind}
~\newline
{\bf Input:} 

\quad - Data consisting of the collection of 

\quad ~~~observations $X$

\quad - Training patterns $P_1,\ldots,P_n$

\quad - Seed $S \subset X$ of a new potential pattern

\quad ~~~to be discovered.
\newline\newline
{\bf Output:} New pattern $\hat{P}$.
\newline\newline
Initialize $\hat{P}=S$. \vspace*{5pt}\\
Step 1. Train the pattern discovery algorithm on the known patterns $Q=\{P_1,\dots,P_n\}$ , as follows:
\(
\beta^{*} = \min_{\beta} \hR_{\alpha,n}(f_{\beta}) = \min_{\beta} n^{-1} \sum_{j=1}^{n} \mcL_{\alpha}(f_{\beta}, P_{j}, X_{Q})
\)
using the definitions from Section \ref{sec:individual} to define the loss function and selector function. Let
$f^{*} = f_{\beta^{*}}.$
\newline\newline
Step 2. Find new pattern.
\newline\newline
{\bf while} $f^{*}(\hat{P})>\theta$ {\bf do}

\quad Compute the best observation to add to the 

\quad set $\hat{P}$:
      \(\hat{x}\in \textrm{argmax}_x f^{*}(\hat{P}\cup x).\)

\quad{\bf if} $f^{*}(\hat{P}\cup \hat{x})>\theta$ {\bf then}

\quad\qquad the pattern has a sufficiently high score, 

\quad\qquad and we should add $\hat{x}$ to the pattern:
       \(\hat{P}\leftarrow \hat{P}\cup \hat{x}.\)
  \vspace*{5pt}
Return $\hat{P}$. \\
\rule{.45\textwidth}{.1pt}
\end{algorithm}

This algorithm has the advantage of being computationally tractable, and is directly motivated by our choice of selector function. In order to select the optimal parameter $\beta^{*}$, we need only consider subsets of the true patterns, along with an additional observation.
Also, for growing new patterns, the function $f^{*}$ was specifically trained to be able to distinguish observations that belong in the pattern from those that do not belong, which is ideal for this method.

\subsection{Application to Crime Pattern Detection} \label{sec:crimes}
An algorithm that is extremely similar to the one provided above was used by \citet{Wang:2013} to detect crime series in the city of Cambridge, MA. In that application:
\begin{itemize}
\item $X$ is a set of crimes, namely housebreaks, that happened between 1996 and 2007 in Cambridge, MA. Many details of each crime are available, including the date, time, day of the week, location, type of premise (apartment, house), location of entry (window, back door), means of entry (pried, cut screen), whether the dwelling was ransacked, etc. 
\item $P_1,\ldots,P_n$ is a database of known crime patterns provided by the Cambridge police department that had been curated by their Crime Analysis Unit over the decade 1996-2007. Crimes in pattern $P_i$ were all hypothesized to have been committed by the same individual or group (they are ``crime series").
\item $f(P)$, $P \in \SX$, is a nonlinear function of the details of crimes within the pattern,  called ``pattern-crime similarity," parameterized by a vector $\lambda$. In particular, within function $f$ is a linear combination of similarity measures between crimes, where $\lambda$ are the linear coefficients. For instance, if $j$ is the coefficient for location, and the value of the learned $\lambda_j$ is large, it means that location is an important factor in determining whether a set of crimes is indeed part of a crime series.
\item A loss function that is similar to (but slightly different than) the one provided in Section \ref{sec:individual} was used to train the algorithm on past patterns $P_1,\ldots,P_n$ along with the rest of the crimes $\mathcal{X}$ to determine values for vector $\lambda$.
\item A threshold similar to $\theta$ was used to determine when to stop growing the crime pattern. In particular, when the series becomes less cohesive after adding more crimes, the series is considered to be complete.
\end{itemize}
The algorithm of \citet{Wang:2013} has been successful in being able to detect patterns of crime in Cambridge, and in a blind test with Cambridge crime analysts, it has been able to locate 9 crimes that belong in patterns that were not previously identified as such, and it was able to exclude 8 crimes that analysts previously thought were part of a pattern (they now agree that these crimes are not part of a pattern). In one case, the exclusion of a crime from a pattern helped to narrow the suspect description down to one possible race and gender (white male).


\subsection{Application to Set Completion and ``Growing a List" in Information Retrieval} \label{sec:list-grow}
Another algorithm similar to Algorithm \ref{Alg:Ind} was used for the problem of set completion in information retrieval. A ``set completion engine" is a next generation search engine. It takes a few seed examples, of almost anything, and simply aims to produce more of them. For instance, a search starting with seed ``Boston Harborfest" and ``South Boston Street Festival" should yield a list of more large annual events in Boston. The algorithm of \citet{LethamRuHe13} for growing a list of items from a seed uses an algorithm similar to Algorithm \ref{Alg:Ind} in that at each iteration, a new item is added to the set. Here:
\begin{itemize} 
\item $X$ is a set of all terms and phrases found on the internet.
\item $P_1,\ldots,P_n$ is a set of gold standard completed sets, such as the ``List of $\ldots$" articles on Wikipedia used for experiments of \citet{LethamRuHe13}.
\item $f(P)$, $P \in \SX$, is a linear combination of similarities between terms, with coefficients chosen for the Bayesian Sets algorithm of \citet{GhahramaniHeller}. This algorithm is unsupervised, meaning that the training step in Algorithm \ref{Alg:Ind} is replaced with a Bayesian prior. 
It would not be difficult to design a supervised algorithm that learns the prior hyper-parameters of Bayesian Sets, rather than having the user choose them. In the experiments of \citet{LethamRuHe13}, the parameters were chosen using a heuristic. The terms that are combined are indicator variables of the internet domains where items can be found.
\end{itemize}
In the case of growing a list, \citet{LethamRuHe13} showed that as long as the feature space and score $f$ are constructed correctly, the results coming from this algorithm are accurate enough to be used in practice, and are substantially more accurate than other methods currently in use for set completion, including Boo!Wa!\footnote{www.boowa.com} and Google Sets.\footnote{Google Sets is available through Google Spreadsheet.} 

The present work thus provides theoretical foundations for the methodologies used by \citet{Wang:2013} and \citet{LethamRuHe13} that we discussed in the Sections \ref{sec:crimes} and \ref{sec:list-grow}.

\appendix

\section{Proof of Theorem 3.1}
Recall the statement of Theorem 3.1:
\bthmn
 Let $Q_{1},\dots,Q_{n} \overset{i.i.d.}{\sim} \mcD$. Then for any positive integer $n$ and any $0 < \delta < 1$, with probability $1 - \delta$ over samples of length $n$, every $f \in \mcF$ satisfies
\begin{align}
L_{n}(f) &\le \bbE  \int 24\sqrt{\frac{\log \mcN(\tilde\mcG, d_{n}, \eps)}{n}}\,d\eps + \sqrt{\frac{8 \ln (2/\delta)}{n}}. \label{eq:l2-bound-g}
\end{align}
\ethmn

\bprf
Let 
\beq \hat\mcR_{n}(\mcF) := \bbE_{\veps}\bigg[\sup_{f \in \mcF} \big| n^{-1}\textstyle\sum_{i=1}^{n} \veps_{i} f(X_{i})\big| \bigg| \{X_{i}\}\bigg], \eeq
where the $\veps_{i}$ are independent uniform $\{\pm 1\}$-valued random variables. Define the {\em Rademacher complexity} to be $\mcR_{n}(\mcF) := \bbE \hat\mcR_{n}(\mcF)$. We will need the following Rademacher complexity-based risk bound.
\bthm[\cite{Bartlett:2002}] \label{thm:rademacher-bound}
Let $Q_{1},\dots,Q_{n} \overset{i.i.d.}{\sim} \mcD$. Then for any positive integer $n$ and any $0 < \delta < 1$, with probability $1 - \delta$ over samples of length $n$, every $f \in \mcF$ satisfies
\beq L_{n}(f) \le 2\mcR_{n}(\tilde\mcG) + \sqrt{\frac{8 \ln (2/\delta)}{n}}. \eeq
\ethm

Recall that $\mcG$ is the loss class and $\tilde\mcG$ the offset loss class. We can relate the Rademacher complexity term in the preceding theorem to the empirical covering number of $\mcF$. This can be done via Dudley's entropy bound, which we state just for case of the Rademacher process 
\beq \hat\mcR_{n} (g) = n^{-1/2}\sum_{i=1}^{n} \veps_{i} g(Q_{i}). \eeq

\bthm[Dudley's entropy bound; \cite{Mendelson:2002}] 
For the Rademacher process $\hat\mcR_{n}$ defined above,
\beq \bbE \sup_{g \in \mcG} \hat\mcR_{n}(g) \le \int 12 \sqrt{\log \mcN(\mcF, d_{n}, \eps)}\,d\eps. \eeq
\ethm

Since $\mcR_{n}(\mcG) = \bbE\sup_{g \in \mcG} n^{-1/2}\hat\mcR_{n}(g)$, combining the two theorems gives result.
\eprf

\section{Proof of Theorem 3.2}
Recall the statement of Theorem 3.2: 
\bthmn
Under the same hypotheses as Theorem 3.1,
\begin{align*}
L_{n}(f) &\le \bbE \int 24 \sqrt{\frac{\log \mcN(\mcF, \ell_{n}, \eps)}{n}}\,d\eps +  \sqrt{\frac{8 \ln (2/\delta)}{n}}. 
\end{align*}
\ethmn
\bprf
Let  $I_{Q}(X_Q, U) := \mathbb I(U \subset P \in Q)$. Then for $g \in \tilde\mcG$ we have
\beqa
g(Q) &=& \mcL_{S}(f; Q) - \mcL_{S}(0; Q) = \ell_{Q}(f, I_{Q}) - \ell_{Q}(0, I_{Q}), \textrm{ so }
\eeqa
\begin{align*}
 d_{n}(g_{1},g_{2}) &= \left[n^{-1}\sum_{i=1}^{n} (g_{1}(Q_{i}) - g_{2}(Q_{i}))^{2}\right]^{1/2} \\
 &=  \left[n^{-1}\sum_{i=1}^{n} (\ell_{Q_{i}}(f_{1}, I_{Q_{i}}) - \ell_{Q_{i}}(f_{2}, I_{Q_{i}}))^{2}\right]^{1/2} \\
&\le  \left[n^{-1}\sum_{i=1}^{n} \ell_{Q_{i}}^{2}(f_{1}, f_{2})\right]^{1/2} = \ell_{n}(f_{1},f_{2}). 
\end{align*}
This inequality implies that if the $\eps$-ball centered at $f$ (with metric $\ell_{n}$) contains $f'$ then the $\eps$-ball centered at $g$ (with metric $d_{n}$) contains $g'$. Hence $\mcN(\mcG, d_{n}, \eps) \le \mcN(\mcF, \ell_{n}, \eps)$, which together with Theorem 3.1 gives the result.
\eprf

\section{Proof of Lemma 5.1}
Recall the statement of Lemma 5.1:
\blemn 
Assuming $|P_{i}| \le B$ almost surely, then if one $P_{i}$ changes, the value of $\hR_{\alpha,n}(f)$ changes by at most $B_{\alpha}/n = \left(1 + 2(1-\alpha)B\right)/n$.  
\elemn
\bprf Let $\Delta \mcL_{\alpha}^{j,i}$ denote the maximum possible change in $\mcL_{\alpha}(f, P_{j}, X_{Q})$ due to a change in $P_{i}$:
\beq \Delta \mcL_{\alpha}^{j,i} = \sup_{P_{i}'} |\mcL_{\alpha}(f,P_{j},X_{Q}) - \mcL_{\alpha}(f,P_{j},X_{Q'})|, \eeq
where $Q' = \{ P_{1},P_{2},\dots,P_{i-1},P_{i}',P_{i+1},\dots,P_{n}\}$. 
Recall that $\mcL_{\alpha}(f; P_{j}, X_{Q})$ consists of a sum of losses over $\mathscr S(P_{j})$ weighted by $\alpha/Z_{P_{j}}$ and a sum of losses over $S_{-}(P_{j},X_{Q})$ weighted by $(1-\alpha)/Z_{P_{j},X_{Q}}$.

For $j \ne i$, the sum
\beq \frac{\alpha}{Z_{P_{j}}}\sum_{U \in \mathscr S(P_{j})} \ell_{+}(f, U)  \eeq
from $\mcL_{\alpha}(f, P_{j}, X_{Q})$ remains constant when $P_{i}$ changes since it does not depend on $X_{Q}$. The second normalizing constant is 
\beq Z_{P_{j},X_{Q}} = (|X_{Q}| - |P_{j}|)(2^{|P_{j}|}-1), \eeq
which is the number of nontrivial subsets of $P_j$ combined with one element that is not from $P_j$.

Write 
\(
Z &= Z_{P_{j},X_{Q}} \\
Z' &= Z_{P_{j},X_{Q'}} \\
Y &= \sum_{U \in S_{-}(P_{j}, X_{Q})} \ell_{-}(f, U) \\
\Delta Y &= Y - \sum_{U \in S_{-}(P_{j}, X_{Q'})} \ell_{-}(f, U)
\)
so we have 
\beqa
\left|\frac{Y}{Z} - \frac{Y-\Delta Y}{Z'}\right| &=& \frac{|Z'Y - Z(Y-\Delta Y)|}{ZZ'} \\
&\le& \frac{Y|Z' - Z|}{ZZ'}  + \frac{|\Delta Y|}{Z'} \\
&\le& \frac{|Z' - Z| + |\Delta Y|}{Z'},
\eeqa
where we used that $Y/Z\leq 1$.
At most $|P_{i}|(2^{|P_{j}|}-1)$ terms in the sum 
\beq \sum_{U \in S_{-}(P_{j}, X_{Q})} \ell_{-}(f, U) \eeq
can change value when $P_{i}$ changes. This is the number of subsets of $P_j$ combined with one element of $P_i$.
Since $\ell_{-} \in [0,1]$ and $|P_{i}| \le B$, we therefore have 
\(
\Delta Y \le |P_{i}|(2^{|P_{j}|}-1) \le B(2^{|P_{j}|}-1).
\)
Also, 
\beq |Z-Z'| = (|P_{i}|-|P_{i}'|)(2^{|P_{j}|}-1) \le B(2^{|P_{j}|}-1) \eeq
and
\beq Z' = (|X_{Q'}| - |P_{j}|)(2^{|P_{j}|}-1). \eeq
Combining these we get (for $j \ne i$)
\(
\Delta \mcL_{\alpha}^{j,i} 
&\le (1-\alpha) \left|\frac{Y}{Z} - \frac{Y-\Delta Y}{Z'}\right| \\
&\le \frac{2(1-\alpha)B(2^{|P_{j}|}-1)}{(|X_{Q'}| - |P_{j}|)(2^{|P_{j}|}-1)} \\
&\le \frac{2(1-\alpha)B}{n - 1},
\)
where we have used the fact that $|X_{Q'}| - |P_{j}| \ge n - 1$, since there is at least one element in each pattern.
%
We also have the trivial bound that $\Delta \mcL_{\alpha}^{i,i} \le 1$ because $\mcL_{\alpha} \in [0,1]$. Letting $\hR_{\alpha,n}'(f)$ denote $\hR_{\alpha,n}(f)$ when $P_{i}$ is replaced by $P_{i}'$, we have 
\(
\lefteqn{\sup_{P_{i}'}|\hR_{\alpha,n}(f) - \hR_{\alpha,n}'(f)| \le} \\
& n^{-1} \sum_{j} \Delta\mcL_{\alpha}^{j,i} \le n^{-1} \left(1 + \sum_{j \ne i} \Delta\mcL_{\alpha}^{j,i}\right) \\
&\le n^{-1}\bigg(1 + (n-1)\frac{2(1-\alpha)B)}{n - 1}\bigg) \\
&= \frac{1 + 2(1-\alpha)B}{n} = \frac{B_{\alpha}}{n}.
\)
\eprf

\section{Proof of Theorem 5.1}
Recall the statement of Theorem 5.1:
\bthmn 
Let $P_{1},\dots,P_{n} \overset{i.i.d.}{\sim} \mcP$ and assume $|P_{i}| \le B$ almost surely. Then for any positive integer $n$ and any $0 < \delta < 1$, with probability $1 - \delta$ over samples of length $n$, every $f \in \mcF$ satisfies
\beqn L_{\alpha,n}(f) \le 2\mcQ_{n}(\tilde\mcL_{\alpha}, \mcF) + \sqrt{\frac{8B_{\alpha}^{2} \ln (2/\delta)}{n}}, \eeqn
where $B_{\alpha} = 1 + 2(1-\alpha)B$ and $\tilde\mcL_{\alpha}(f; \cdot, \cdot) := \mcL_{\alpha}(f; \cdot, \cdot) - \mcL_{\alpha}(0; \cdot, \cdot)$ is the shifted loss. 
\ethmn
\bprf 
We have
\[
R_{\alpha,n}(f) &\le  \hR_{\alpha,n}(f) + \sup_{f' \in \mcF} \big(R_{\alpha,n}(f') -  \hR_{\alpha,n}(f')\big) \nonumber \\
&=  \hR_{\alpha,n}(f)  +  R_{\alpha,n}(0) -  \hR_{\alpha,n}(0) \nonumber \\
&\quad + \sup_{f' \in \mcF} \big\{R_{\alpha,n}(f') - \hR_{\alpha,n}(f') \nonumber \\
&\qquad\qquad\quad R_{\alpha,n}(0) - \hR_{\alpha,n}(0) \big\} \nonumber \\ 
&=  \hR_{\alpha,n}(f) + R_{\alpha,n}(0) -  \hR_{\alpha,n}(0) \nonumber \\
&\quad +\sup_{f' \in \mcF} \big\{\bbE \tilde R_{\alpha,n}(f') -  \tilde R_{\alpha,n}(f')\big\} ,  \label{eq:bound-me}
\]
where $\tilde R_{\alpha,n}(f') := n^{-1} \sum_{j=1}^{n} \tilde\mcL_{\alpha}(f', P_{j}, X_{Q})$. 

We now wish to bound the final two terms of \eqref{eq:bound-me}. First consider the term $\sup_{f' \in \mcF} \big(\bbE \tilde R_{\alpha,n}(f') -  \tilde R_{\alpha,n}(f')\big)$. Note that $\bbE \tilde R_{\alpha,n}(f')$ is a constant. On the other hand, if one $P_{i}$ changes, $\tilde R_{\alpha,n}(f')$ can change by at most $2B_{\alpha}/n$ since $\tilde R_{\alpha,n}(f') = \hR_{\alpha,n}(f) - \hR_{\alpha,n}(0)$ and each of these $\hR_{\alpha,n}(\cdot)$ terms can change by at most $B_{\alpha}/n$ by Lemma 5.1. Now applying McDiarmid's inequality with $F = \sup_{f' \in \mcF} \big(\bbE \tilde R_{\alpha,n}(f') -  \tilde R_{\alpha,n}(f')\big)$ and $c = 2B_{\alpha}/n$, we have that with probability at least $1 - \delta/2$, 
\[
\lefteqn{\sup_{f' \in \mcF} \big(\bbE \tilde R_{\alpha,n}(f') -  \tilde R_{\alpha,n}(f')\big)} \label{eq:mcdiarmid-application} \\
&\le \bbE \sup_{f' \in \mcF} \big(\bbE \tilde R_{\alpha,n}(f') - \tilde R_{\alpha,n}(f')\big) + \sqrt{2 B_{\alpha}^{2}\ln(2/\delta)/n}. \nonumber 
\]
An essentially identical argument applies to bounding $R_{\alpha,n}(0) -  \hR_{\alpha,n}(0)$ by noting that $R_{\alpha,n}(0)$ is a constant and that $\bbE[R_{\alpha,n}(0) -  \hR_{\alpha,n}(0)] = 0$, so with probability at least $1 - \delta/2$, 
\beqn
R_{\alpha,n}(0) -  \hR_{\alpha,n}(0) \le \sqrt{2B_{\alpha}^{2}\ln(2/\delta)/n}.
\eeqn
Combining the previous two bounds with \eqref{eq:bound-me} gives, with probability at least $1-\delta$,
\begin{align}
R_{\alpha,n}(f) 
&\le  \hR_{\alpha,n}(f) + \sqrt{\frac{8B_{\alpha}^{2} \ln (2/\delta)}{n}} \\
&\quad+ \bbE \sup_{f' \in \mcF} \big(\bbE\tilde R_{\alpha,n}(f') - \tilde R_{\alpha,n}(f')\big) \nonumber.
\end{align}
To complete the proof, let $P_{1}',\dots,P_{n}' \overset{i.i.d.}{\sim} \mcP$ and let $Q' = \{ P_{1}', \dots, P_{n}'\}$. Now, writing what the expectations are with respect to for clarity, we have
\begin{align*}
\lefteqn{\bbE_{Q} \sup_{f' \in \mcF} \big(\bbE_{Q'}\tilde R_{\alpha,n}(f') - \tilde R_{\alpha,n}(f')\big)} \\
&= \bbE_{Q} \sup_{f' \in \mcF} \bbE_{Q'} \Bigg[n^{-1} \sum_{j=1}^{n} \tilde\mcL_{\alpha}(f', P_{j}', X_{Q'}) - \tilde R_{\alpha,n}(f') \bigg| Q\Bigg] \\
&\le \bbE_{Q,Q'} \sup_{f' \in \mcF}\left[n^{-1} \sum_{j=1}^{n} \tilde\mcL_{\alpha}(f', P_{j}', X_{Q'}) - \tilde R_{\alpha,n}(f')\right]  \\
&= \bbE_{Q,Q',\veps} \sup_{f' \in \mcF}\Bigg[n^{-1} \sum_{j=1}^{n} \veps_{i}\big(\tilde\mcL_{\alpha}(f', P_{j}', X_{Q'}) - \tilde\mcL_{\alpha}(f', P_{j}, X_{Q})\big)\Bigg]  \\
&\le  2 \bbE_{Q,\veps} \sup_{f' \in \mcF}\left[n^{-1} \sum_{j=1}^{n} \veps_{i}\tilde\mcL_{\alpha}(f', P_{j}, X_{Q})\right]  \\
&\le 2 \mcQ_{n}(\tilde\mcL_{\alpha}, \mcF). 
\end{align*}
The first line follows from the definition of $\tilde R_{\alpha,n}(f')$. The first inequality follows from Jensen's inequality applied to $\sup$. The second equality follows by symmetry. The second inequality follows since symmetry permits the difference in each pair of $\tilde\mcL_{\alpha}$ terms can be bounded by twice the (worst) of one term. The last inequality relies on Jensen's inequality applied to $|\cdot|$ and the fact that $|\sup \cdot| \le \sup |\cdot|$. For more details about this type of proof technique, see \cite{Bartlett:2002}.
\eprf

\section{Proof of Lemma 5.2}
Recall the statement of Lemma 5.2:
\blemn
If $|P_{i}|$ has geometric tails, then the expected number of observations in a pattern is at most 
\beqn B_{\lambda,C} :=  B_{0} + \frac{C\lambda^{B_{0}+1}}{1 - \lambda}(B_{0} + 1/(1 -\lambda)). \eeqn
\elemn
\bprf
Since $\Pr[|P| \ge B] \le C\lambda^{B}$ for $B \ge B_{0}$, $P_{B} := \Pr[|P| = B] \le C\lambda^{B}$ for $B \ge B_{0}$. Note that 
\beqa
\lambda \D{}{\lambda}\sum_{B=B_{0}}^{\infty} \lambda^{B} &=& \lambda \D{}{\lambda}\frac{\lambda^{B_{0}}}{1 -\lambda} \\
\sum_{B=B_{0}}^{\infty} B\lambda^{B} &=& \lambda\frac{B_{0}\lambda^{B_{0}-1}(1-\lambda) + \lambda^{B_{0}}}{(1 -\lambda)^{2}} \\
&=& \lambda^{B_{0}}\frac{B_{0}(1-\lambda) + \lambda}{(1 -\lambda)^{2}}.
\eeqa
Hence, 
\beqa
\bbE[|P|] &\le& B_{0} + \sum_{B = B_{0}+1}^{\infty} B P_{B} \\
&\le& B_{0}(1 - C\lambda^{B_{0}}) + C \sum_{B=B_{0}}^{\infty} B \lambda^{B}  \\
&=& B_{0}(1 - C\lambda^{B_{0}}) + C\lambda^{B_{0}}\frac{B_{0}(1-\lambda) + \lambda}{(1 -\lambda)^{2}} \\
&=& B_{0} + \frac{C\lambda^{B_{0}+1}}{1 - \lambda}(B_{0} + 1/(1 -\lambda)).
\eeqa
Note the change in the start of the summation between the first and second line. 
\eprf

\section*{Acknowledgements}

Thanks to Dylan Kotliar and Yakir Reshef for helpful discussions regarding personalized medicine and cancer genomics and to Peter Krafft for helpful comments. Funding provided by MIT Lincoln Laboratory and NSF-CAREER IIS-1053407. Supplementary material is available at the first author's personal website. 

\bibliographystyle{plainnat}
\bibliography{mybib}

\end{document}